\documentclass{article}
\pdfoutput=1

     \usepackage[nonatbib, preprint]{neurips_2021}

\usepackage[numbers]{natbib}

\usepackage[utf8]{inputenc} 
\usepackage[T1]{fontenc}    
\usepackage{hyperref}       
\usepackage{url}            
\usepackage{booktabs}       
\usepackage{amsfonts}       
\usepackage{nicefrac}       
\usepackage{microtype}      

\usepackage{dsfont}

\usepackage{amsmath}
\DeclareMathOperator{\argmax}{argmax}

\usepackage{algorithm}
\usepackage{algorithmic}
\usepackage{subcaption}
\usepackage{xcolor}
\usepackage{microtype}
\usepackage{graphicx}
\usepackage{booktabs,amsmath} 

\usepackage{microtype}
\usepackage{graphicx}
\usepackage{booktabs,amsmath} 

\title{Unsupervised Neural Hidden Markov Models with~a~Continuous~latent~state~space}

\author{%
  Firas JARBOUI \\
  ANEO \\
  Centre Borelli - ENS Paris-saclay  \\
  \texttt{firasjarboui@gmail.com} \\
  \And
  Vianey PERCHET \\
  Criteo AI Lab \\
  Crest, ENSAE \\
  \texttt{vianney.perchet@normalesup.org} \\
}

\begin{document}

\maketitle

\begin{abstract}
  We introduce a new procedure to neuralize unsupervised Hidden Markov Models in the continuous case. This provides higher flexibility to solve problems with underlying latent variables. This approach is evaluated on both synthetic and real data. On top of generating likely model parameters with comparable performances to off-the-shelf neural architecture (LSTMs, GRUs,..), the obtained results are easily interpretable.
\end{abstract}

\section{Introduction}
    Probabilistic Graphical Models (PGM) are  standard techniques to discover highly interpretable hidden structure in data \citep{PGM_inter, XAI}. Even with sequential data, off-the-shelf PGMs such as Hidden Markov Models (HMMs) are able to uncover latent patterns, as assessed by recent breakthroughs in Natural Language Processing \citep{HMM_NLP0, HMM_NLP1}, Speech Recognition \citep{HMM_speech0, HMM_speech1} and Reinforcement Learning \citep{HMM_RL0, HMM_RL1, HMM_RL2}. 
    
    Using flexible parametric probability distributions to model the dynamics of PGMS yields better and more accurate results \citep{PGM_dist}. Given that deep neural networks are universal approximators that can eventually learn highly expressive non-convex functions \citep{MLP}, PGMs were recently and successfully interfaced with Artificial Neural Networks (ANNs) \citep{PGM_NN0, PGM_NN1}. This motivates  to further  bridge neural networks with HMMs, as already attempted back in the 90s \citep{NNxHMM}. Indeed, hybrid HMM/ANN models with better accuracy  have been created by extracting features using neural networks : typical examples include word embeddings for unsupervised word alignment \citep{embedding}, convolutional neural networks for speech recognition \citep{conv}, etc.
    %
    
    A generic approach \citep{neuralHMM} to parameter estimation in finite latent space neural HMMs, suggests to model the parameters  with neural networks. In that case, gradients are back-propagated with respect to the neural network parameters during the maximisation step (M) of the Baum-Welch algorithm (the Viterbi algorithm is used in the E-step). Our main objective is to introduce a new procedure for the continuous latent state space case with a similar flavor. 
    
    More precisely, we will consider HMMs with continuous latent spaces as they encompass a very large class of non-parametric models well adapted to sequential processes. However, they were recently outperformed by recurrent neural networks (RNNs), which led to some decreasing interest \citep{HMMvsLSTM}. On the other hand, they are  highly interpretable, a key property not necessarily shared with RNN. Our main contribution is to advocate that using neural networks to estimate the parameters of the transition and emission probabilities of an HMM improves its performance up to be comparable to  RNNs, while retaining the interpretability of the latent variables, therefore reaching the best of both worlds: \textbf{performances} and \textbf{interpretability}. 
    
    In details, we will investigate the theoretical framework of SMC methods to construct an optimisation algorithm that maximises the log-likelihood of the observations with respect to the model neural network parameters. Driven by  the successes of stochastic gradient descent, we also provide a stochastic and more efficient variant that preserves its experimental effectiveness. 
    
    Our methods  are evaluated both on simulated and real data. The toy example is the simulation of an HMM and, without exploiting prior knowledge of the environment, our algorithm solves the parameters estimation problem. The two real-life applications use data from tourism and music. Their analysis showcase the simplicity of the implementation of our methods and, more importantly, strongly highlight the interpretability of the obtained results.
\section{Neural Hidden Markov Model}
\label{sec:neuralHMM}
    In order to define ANN-based HMM, we consider two neural networks $f_\theta$ and $g_\theta$. To alleviate notations, we use $\theta$ to denote the parameters of both ANN, even though the two networks do not necessarily share parameters. Without loss of generality, we consider a latent space $\mathcal{X} = \mathds{R}^{d_h}$ and an observation space $\mathcal{Y}=\mathds{R}^{d_o}$. Otherwise,  an additional embedding layer would do the trick. The neural-equivalent version of HMM models (c.f. appendix \ref{sec:Background} for a review of HMM models) writes as:
    \begin{equation}
        \left .
        \begin{array}{ccc}
            x_0  \sim  \mu(x) ; &
            y_t  \sim  p_{g_\theta(x_t)}(y) ; &
            x_{t+1}  \sim  q_{f_\theta(x_t)}(x)
        \end{array}
        \right .
        \label{HMM_neural}
    \end{equation}{}
    In practical applications, the objectives are to optimise the model parameters for multiple independent trajectories $(y^j_{0:T})_{j=0}^K$. Thus, the log-likelihood of the observations under a given $\theta$ can be written as: 
    \begin{equation}
        \left.
        \begin{array}{lll}
            \mathcal{L}(\theta, (y^j_{0:T})_{j=0}^K) =\sum_j \mathcal{L}_j(\theta, y^j_{0:T}) ; &
            \mathcal{L}_j(\theta, y^j_{0:T}) = \log\mathrm{P}(y^j_{0:T}|\theta) 
        \end{array}
        \right .
    \end{equation}{}
     Equation \eqref{HMM_neural} introduces non-linearity in the system, hence SMC approaches are well suited to optimise $\mathcal{L}$. Inspired by EM algorithms (see Appendix \ref{sec:ParticleFilters}),  the particle filter from Algorithm \ref{bootstrap} can be used to construct an asymptotically unbiased estimator of the $\mathcal{Q}$ function.
    
    Unfortunately, the maximiser of the following function $\hat{\mathcal{Q}}$, that approximates $\mathcal{Q}$, does not have an explicit form in the case of the neural HMM (in Equation \eqref{HMM_neural}), because the emission and transition probabilities are no longer within the exponential family. This function $\hat{\mathcal{Q}}$ is defined by 
    \begin{equation}
        \left .
        \begin{array}{r} \hat{\mathcal{Q}}(\theta, \theta_k) = \sum_{i,j,t} W^{i,j}_T \big[\log\big( p_{g_\theta(\hat{X}^{i,j}_t)}(y_t^j) \big) 
            + \log\big( q_{f_\theta(\hat{X}^{i,j}_{t-1})}(\hat{X}^{i,j}_t )  \big)\big],
        \end{array}
        \right.
       \label{Q_approx_nn}
    \end{equation}
    \noindent where $\hat{X}^{i,j}_t$ is the $i^\textit{th}$ particle generated at time $t$ for the $j^\textit{th}$ trajectory using an SMC method and $y_t^j$ is the associated observation. However, it is possible, using  Equation \eqref{EM_strong}, to prove that a candidate parameter $\theta_{k+1}$ must satisfy: 
    \begin{equation}
       \hat{\mathcal{Q}}(\theta_{k+1}, \theta_k) \geq \hat{\mathcal{Q}}(\theta_k, \theta_k)
    \end{equation}{}
    As long as the probability distribution functions associated with the emission and transition probabilities are differentiable, the gradient of $\hat{\mathcal{Q}}$ is well defined. Moreover, functions of the form of Equation \eqref{Q_approx_nn} are well suited for stochastic gradient-based (SGD) optimisation. In fact, the loss function $l$ introduced in Equation \eqref{loss} for a given sample $s = (y_t^j, \hat{X}^{i,j}_t, \hat{X}^{i,j}_{t-1})$ can be used to construct a new parameter $\theta_{k+1}$, improving the  likelihood:
    \begin{equation}
        \left .
        \begin{array}{r} l_{\theta_k}(s)(\theta) =  W^{i,j}_T \big[\log\big( p_{g_\theta(\hat{X}^{i,j}_t)}(y_t^j) \big) \\
            + \log\big( q_{f_\theta(\hat{X}^{i,j}_{t-1})}(\hat{X}^{i,j}_t )  \big)\big]
        \end{array}
        \right .\ .
        \label{loss}
    \end{equation}{}
    Optimising the log-likelihood with EM requires the evaluation of the posterior distribution approximation through the estimation of $\big[(X^{i,j}_{0:T}), (W^{i,j}_T) \big]_{i,j=0}^{N,K}$ in the E-step, and maximising the $\hat{\mathcal{Q}}$ function in the M-step. Equation \eqref{Q_neural_HMM_} provides an iterative algorithm that solves neural HMMs:
    \begin{equation}
        \left \{
        \begin{array}{l} (\hat{X}^{i,j}_t)_{i,j,t}, (W^{i,j}_T)_{i,j} = \textit{SMC}(\theta_k, (Y^j_t)_{j,t}) \\
            \theta_{k+1} = \textit{SGD}(l, \theta_k, (\hat{X}^{i,j}_t)_{i,j,t}, (W^{i,j}_T)_{i,j}, (Y^j_t)_{j,t})
        \end{array}
        \right .
     \label{Q_neural_HMM_}
    \end{equation}
    
    \subsection{Reducing computational complexity}
        A drawback of the EM algorithm is that the (E) step is computationally expensive. The main reason is that Equation~\eqref{Q_neural_HMM_} requires to run the particle filter over the full data set. When the available trajectories share a lot of similarities, we suggest that this is not necessary. An alternative is to randomly sample a batch of trajectories on which we run the optimisation. This obviously leads to faster convergence. The downside is that  the same performance might not be reached; however, this is highly mitigated with a  few steps at the end of training where we use the full data set in order to fine tune the parameters.
        We will discuss this further in the following section. 
    
\section{Experiments}
\label{sec:experience}
\begin{figure}
            \centering
                \begin{subfigure}{0.45\linewidth}
                    \includegraphics[width=\linewidth]{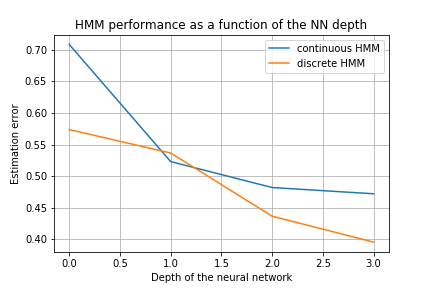}
                    \caption{Impact of the depth}
                    \label{depth}
                \end{subfigure}
                \begin{subfigure}{0.45\linewidth}
                    \includegraphics[width=\linewidth]{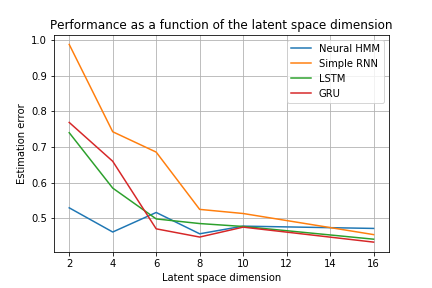}
                    \caption{Impact of the latent space dimension}
                    \label{RNNvsHMM}
                \end{subfigure}
                \begin{subfigure}{0.45\linewidth}
                    \includegraphics[width=\linewidth]{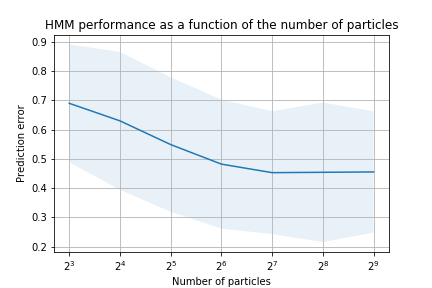} \caption{Impact of the number of particles}
                    \label{N_particles}
                \end{subfigure}
                \begin{subfigure}{0.45\linewidth}
                    \includegraphics[width=\linewidth]{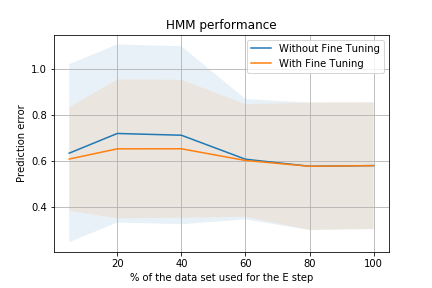}
                    \caption{Impact of the \% of the data set used}
                    \label{randomize}
                \end{subfigure}
                \caption{Performance analysis of neural HMM as a function of different parameters}
                \label{perf_MCP}
            \end{figure}

    We now provide experimental results to assess the relevancy and efficiency of our approach. In the synthetic example (where data are generated by some HMM), the environment is a set of target positions in a d-dimensional space, and the observations are the trajectories of a particle bouncing around them (a description of the data generating process is provided in Appendix \ref{sec:generatif process}). We compare the performance of our parameter inference scheme with classical approaches to train HMMs as well as off the shelf neural network architectures (simple RNNs, LSTMs, GRUs,..). This establishes the ability of neural HMMs to provide comparable results to RNNs.
    In the real life examples, we use tourists GPS tracks in an open air museum, and guitar chords recordings. Experimental results of those experiments are provided in the Appendix.
    Based on those empirical results, we can establish that neural networks does not affect the ability of HMMs to infer interpretable latent states !
    \paragraph{Depth of the Neural Network}
            We compare the performance of different HMM models on the task of predicting the next observation. In the continuous HMM, the probability distributions used for the emission and transition model are Gaussian. The functions $f_\theta$ and $g_\theta$ are $n$-layer deep neural network used to estimate the mean and variance of the distribution, with $n$ designating the depth of the neural network. The case where $n=0$ is associated to vanilla HMMs where the parameters are learned using Equation \eqref{EM_HMM_exp_family}. For the other depth values, we use Equation \eqref{Q_neural_HMM_}. 
            In  discrete HMMs, the transition probability distribution is a multinomial over $\{1,2,3,4,5\}$. $f_\theta$ is an n-layer neural network predicting the multinomial probabilities. In the case where $n=0$, we use Baum-Welch to learn the parameters, otherwise we use the neural version provided in \cite{neuralHMM}. \\
            In Figure \ref{depth}, we observe that using ANN to learn the HMM parameters yields an improved performance. The obtained results from the continuous HMM case are comparable to the ones obtained from the discrete case. Deeper neural networks provided higher log-likelihoods and perform better on the one step prediction task.
            
        \paragraph{Latent space dimension:}
            Let us now compare neural HMM performance (where the parameters of the emission and transition probabilities  are predicted with a 3-layer  ANNs) to off-the-shelf neural network architecture used for sequence prediction (where a RNN's output is fed to a 3-layer deep ANN). 
            We compare performances for different latent space dimensions. In the HMM case, this corresponds to $d_h$, and in the other cases, it corresponds to the output dimension of the recurrent layer.
            In Figure \ref{RNNvsHMM}, we observe that at higher latent space dimension, all the models have comparable performances. 
            However, at lower dimension, the neural HMM outperforms by a wide margin all the other off-the-shelf methods. This experiment is re-conducted with higher feature dimensions $d$. The obtained results, presented in Table \ref{result_table}, confirm this statement. 
            These results indicate that there is a trade-off between the interpretability of the latent variables and the accuracy of the prediction to be taken into account when choosing an appropriate model. Even though recurrent neural architecture have wildly replaced HMMs in the recent years, we believe neural HMM should mitigate this trend.
            
            \begin{table*}[h]
            \caption{Prediction error for various feature and latent space dimensions.}
            \label{result_table}
            \vskip 0.15in
            \begin{center}
            \begin{small}
            \begin{sc}
            \begin{tabular}{clccccccc}
            \toprule
            Feature dim ($d$) & Models & $d_h = 2$ & $d_h = 3$ & $d_h = 4$ & $d_h = 5$ & $d_h = 10$ & $d_h = 20$  \\
            \midrule
                 & neural HMM & \textbf{.65$\pm$.17}& \textbf{.64$\pm$.17}& \textbf{.66$\pm$.16}& .63$\pm$.16& .65$\pm$.19 & .66$\pm$.18 \\
                 &  HMM       & .92$\pm$.20& .82$\pm$.16& .82$\pm$.18& .76$\pm$.13& .71$\pm$.15 & .71$\pm$.13 \\
            3    &  RNN       & .86$\pm$.06& .83$\pm$.06& .82$\pm$.07& .73$\pm$.10& .64$\pm$.09 & .56$\pm$.10 \\
                 & LSTM       & .71$\pm$.10& .71$\pm$.08& .75$\pm$.09& \textbf{.60$\pm$.09}& .56$\pm$.10 & \textbf{.54$\pm$.10} \\
                 & GRU        & .82$\pm$.06& .82$\pm$.08& .73$\pm$.08& .67$\pm$.08& \textbf{.56$\pm$.0}8 & \textbf{.54$\pm$.09} \\
            \midrule
                 & neural HMM & \textbf{.76$\pm$.13}& \textbf{.77$\pm$.12}& .80$\pm$.13& \textbf{.75$\pm$.14}& .78$\pm$.13 & .78$\pm$.13 \\
                 &  HMM       & .98$\pm$.13& .93$\pm$.11& .87$\pm$.11& .87$\pm$.10& .81$\pm$.12 & .81$\pm$.10 \\
            5    &  RNN       & .98$\pm$.01& .94$\pm$.02& .86$\pm$.06& .83$\pm$.07& .71$\pm$.07 & .65$\pm$.07 \\
                 & LSTM       & .84$\pm$.07& .79$\pm$.07& \textbf{.76$\pm$.08}& \textbf{.76$\pm$.07}& \textbf{.61$\pm$.07} & .61$\pm$.07 \\
                 & GRU        & .87$\pm$.04& .86$\pm$.04& .80$\pm$.05& \textbf{.75$\pm$.05}& .66$\pm$.06 & \textbf{.59$\pm$.07} \\
            \bottomrule
            \end{tabular}
            \end{sc}
            \end{small}
            \end{center}
            \vskip -0.1in
            \end{table*}

        \paragraph{Number of particles:}
            The performance of the algorithm provided in Equation \eqref{Q_neural_HMM_} depends on the accuracy of the $\mathcal{Q}$ function approximation. The performance of the used particle filter increases as the number of particles increases as stated in Equation \eqref{accuracy_PF}. A better approximation allows Equation \eqref{Q_neural_HMM_} to converge to better parameters at the cost of a longer training time. This improvement is however bounded. Increasing the number of particles increases linearly the computational load whereas it improves logarithmically the performance. 
            In Figure \ref{N_particles}, we observe how increasing the number of particles from 8 to 128 improves the performances significantly, whereas increasing the number from 128 to 512 have a marginal impact on the prediction error. The same happens to the average log-likelihood. These experiments were conducted with a neural HMM with 3-layers ANN parameter estimators and with a 5-dimension latent space.
            
        \paragraph{Reducing computational complexity:}
            In order to evaluate the impact of applying iteratively Equation \eqref{Q_neural_HMM_} on randomly sampled subsets instead of the full data set, we use the same setting with higher feature dimension ($d=3$). we compare the prediction error reached after $100$ iteration for different proportions of the data set sampled randomly in the E step. We also evaluate the same error after fine tuning the parameter with $5$ iterations using the whole data set. The conclusion we draw from Figure \ref{randomize} is that randomly sampling trajectories before applying Equation \eqref{Q_neural_HMM_} does not affect a lot the average performance of the obtained parameters. However, the variance of the error is affected. The fine tuning steps can mitigate this issue and result in an improved performance. This procedure is especially useful in settings where an extremely large amount of data is available, such as the guitar chords example that we consider in the last section.
            
\section{Conclusion}
    Neural Hidden Markov Models improve the performance of classical HMMs up to standard RNNs' while providing the same level of latent variable interpretability; thus achieving the best of both worlds. Our main contributions consist in generalising the classical approaches of HMMs training to the neural case and in providing alternatives when the data set is too large. On top of this, implementing neural HMM is a relatively easy task given the available deep learning frameworks. 
    
    The same approach presented in this paper can be adapted to more complex probabilistic graphical models. This is a big step toward reducing gaps between PGMs and RNNs, as it renders HMMs an off-the-shelf tool for practitioner and mitigates the tendency to overlook PGMs in real life applications where explainability is key.

\bibliography{example_paper}
\bibliographystyle{abbrv}

\appendix
\section{Mathematical Background}
\label{sec:Background}
    \subsection{Hidden Markov Models with Continuous latent state space}
        Let us describe formally HMMs. They are defined by some latent state space $\mathcal{X}$ and   observation space $\mathcal{Y}$. Two sequences of observations $(y_t)_{t=0}^T \in \mathcal{Y}$ and  latent states $(x_t)_{t=0}^T \in \mathcal{X}$, where $T\in\mathds{N}$, are generated as follows. The first state $x_0$ is chosen at random according to some underlying distribution $\mu$. Then the first observation $y_0$ is drawn accordingly to 
        $p_\theta(\cdot|x_0)$, a probability distribution over $\mathcal{Y}$. This distribution is called the \textsl{emission probability distribution}. The dynamics of $x_t$ is then controlled by the \textsl{transition probability distribution} $q_\theta(\cdot|x_{t-1})$ as in a standard Markov chain. Formally,  variables satisfy the following equation:
        \begin{equation}
            \left .
            \begin{array}{ccc}
                x_0  \sim  \mu(x); &
                y_t  \sim  p_\theta(y|x_t); &
                x_{t+1}  \sim  q_\theta(x|x_t)
            \end{array}
            \right .
            \label{HMM}
        \end{equation}{}
        The estimation of the model hidden parameters is usually done by the optimisation of the log-likelihood of the observations with respect to the model parameters, i.e.: 
        \begin{equation}
            \theta^* = \argmax_\theta \mathcal{L}(\theta, (y_t)_{t=0}^T) = \argmax_\theta \log \big(\mathrm{P}(y_{0:T} | \theta)\big).
        \end{equation}{}
        As usual, the first step to compute $\mathcal{L}$ is to marginalise this quantity with respect to the latent variables: 
        \begin{equation}
        \mathrm{P}(y_{0:T} | \theta) = \int \mathrm{P}(y_{0:T}, x_{0:T} | \theta)\,dx_{0:T},
        \end{equation}{}
        where the joint distribution can be rewritten into:
        \begin{equation}
        \mathrm{P}(y_{0:T}, x_{0:T} | \theta) = \mu(x_0) \times \prod_{t=0}^T \big[ p_\theta(y_t|x_t) \times q_\theta(x_{t+1}|x_t) \big].
        \label{joint_dist}
        \end{equation}{}
        A popular choice to maximise the log likelihood $\mathcal{L}$ is the class of EM algorithms. Given the initialisation  $\theta_0$, a sequence of estimated parameters is constructed as follows: 
        \begin{equation}
            \left .
            \begin{array}{lcl}
                \theta_{k+1}  =  \argmax_\theta \mathcal{Q}(\theta, \theta_k); \quad
                \mathcal{Q}(\theta, \theta_k)  =  \int \mathrm{P}(x_{:T} | y_{:T}, \theta_k) \log  \mathrm{P}(y_{:T}, x_{:T} | \theta) \,dx_{:T}
            \end{array}{}
            \right .
            \label{EM_HMM}
        \end{equation}{}
        The key property of EM techniques is that the sequence $(\mathcal{L}(\theta_k, (y_t)_{t=0}^T))_k$ is non decreasing. In fact a stronger result \cite{EM1} even states that: 
        \begin{equation}
            \left .
            \begin{array}{c}
                (\mathcal{L}(\theta_, (y_t)_{t=0}^T))_k \geq (\mathcal{L}(\theta_k, (y_t)_{t=0}^T))_k 
                \iff \mathcal{Q}(\theta, \theta_k) \geq \mathcal{Q}(\theta_k, \theta_k)
            \end{array}{}
            \right .
            \label{EM_strong}
        \end{equation}{}
        Moreover, the EM algorithm, is favored to gradient based methods for its numerical stability \cite{EM1}. 
        
    \subsection{Particle Filters for HMMs}
    \label{sec:ParticleFilters}
        In order to optimise the log-likelihood with respect to $\theta$, see Equation \eqref{EM_HMM}, the posterior distribution of the latent states $\mathrm{P}(x_{0:T} | y_{0:T}, \theta)$ must be evaluated. 
        We propose to do this with particle filters, a class of SMC methods. The principle is to generate $N$ latent state candidates, re-sample them according to their likelihood in the model, and use them to generate $N$ candidate latent states for the next time step. By sampling likely trajectories, and weighing them according to their corresponding likelihood, a good approximation of the posterior is constructed. Various implementation of the particle filters can be considered  \cite{aux_filter}.
        
        \begin{algorithm}
        \caption{The bootstrap filter}\label{theta}
        \begin{algorithmic}[1]
        \STATE {\bfseries Procedure:} $\textit{SMC}(\theta, (Y_t)_{t=0}^T) $
        \STATE $X^i_0 \sim \mu_\theta \; \forall i\in[0,N]  $ 
        \STATE $W^i_0 \, \propto \, \mathrm{P}(Y_0|X^i_0, \theta) \times \mu_\theta(X^i_0)$ s.t.  $\sum_i W^i_0 = 1$
        \STATE $\hat{X}^i_0 \sim \sum_i W^i_0 \delta_{X^i_0}$
        \FOR{$t \in  [1,T]$}
        	\STATE $ X^i_t \sim \mathrm{P}(x| \hat{X}^i_{t-1}, \theta)  \; \forall i\in[0,N]  $ 
            \STATE $ W^i_t \, \propto \, \mathrm{P}(Y_t|X^i_t, \theta) \times W^i_{t-1}$ s.t.  $\sum_i W^i_t = 1$
            \STATE $ \hat{X}^i_{0:t} \sim \sum_i W^i_t \delta_{X^i_{0:t}}$
        \ENDFOR
        \STATE {\bfseries Return:} $(\hat{X}^i_{0:T})_{i=0}^N, (W^i_T)_{i=0}^N$
        \end{algorithmic}
        \label{bootstrap}
        \end{algorithm}
        
        The pseudo-code of the bootstrap variation of these filters is presented in Algorithm \ref{bootstrap}. We have denoted by $(X^i_{0:T})_{i=0}^N$ the latent trajectories created by the particle filters, and by  $(W^i_T)_{i=0}^N$, their weights; those two quantities are the keystones of the posterior distribution. Indeed, an asymptotically unbiased estimator is defined by \cite{smoothed_prop, P_filters1}: 
        \begin{equation}
            \hat{\mathrm{P}}(x_{0:T}|y_{0:T},\theta) = \sum_i W^i_T \delta_{\hat{X}^i_{0:T}}(dx_{0:T}).
            \label{estimator}
        \end{equation}{}
        A key property of the particles based estimation of the posterior function is that it can also be used to construct asymptotically unbiased estimates of smooth additive functionals of the following form \cite{smoothed_prop}: 
        \[
            \mathcal{S}_T^\theta = \int \log \big[\prod_{t=0}^T s_t(x_{t-1:t}) \big] \mathrm{P}(x_{0:T} | y_{0:T}, \theta) \, dx_{0:T}
        \]
        Substituting the posterior distribution with its estimate gives an approximation $\hat{\mathcal{S}}_T^\theta$   satisfying \cite{smoothed_prop}:
        \begin{equation}
            \big| \mathds{E}\big[ \hat{\mathcal{S}}_T^\theta \big] - \mathcal{S}_T^\theta \big| \leq F_\theta \frac{T}{N} \; , \; F_\theta \in \mathds{R}^*,
            \label{accuracy_PF}
        \end{equation}
        where the expectation is with respect to the particle filter. Equation \eqref{joint_dist} implies that the $\mathcal{Q}$ function defined in Equation \eqref{EM_HMM} has the same form as $\mathcal{S}_T^\theta$. Notice that this corresponds to the special case where $s_0 = \mu(x_0)$ and  $s_{t+1}(x_{t:t+1}) = p_\theta(y_t|x_t) \times q_\theta(x_{t+1}|x_t)$. 
        As a consequence, an asymptotically unbiased estimator of the $\mathcal{Q}$ function is:
        \begin{equation}
            \hat{\mathcal{Q}}(\theta, \theta_k) = \sum_i W^i_T \log  \mathrm{P}(y_{0:T}, \hat{X}^i_{0:T} | \theta).
            \label{approx_Q}
        \end{equation}{}
        This optimises the log-likelihood $\mathcal{L}$ using an approximation of the EM algorithm presented in Equation \eqref{EM_HMM} by computing at the $k^\textit{th}$ iteration $\argmax_\theta \hat{\mathcal{Q}}(\theta, \theta_k) $. 
    
    \subsection{Implementation}
         In practice, when $p_\theta(y|x_t)$ and $q_\theta(x|x_t)$ are chosen from the exponential family, solving an HMM with the EM algorithm boils down to computing a summary statistic of the $\mathcal{Q}(\theta, \theta_k)$ function using a particle filter. The maximising argument of $\hat{\mathcal{Q}}(\theta, \theta_k)$ can be explicitly computed through a suitable function $\Lambda((Y_t)_{t=0}^T,(\hat{X}^i_{0:T})_{i=0}^N, (W^i_T)_{i=0}^N)$. The standard open source HMM solvers (such as pomegranate and HMMlearn) propose the following algorithm: 
         \begin{equation}
            \left \{
            \begin{array}{lcl}
                (\hat{X}^i_{0:T})_{i=0}^N, (W^i_T)_{i=0}^N = \textit{SMC}(\theta_k, (Y_t)_{t=0}^T)\\
                \theta_{k+1} =  \Lambda((Y_t)_{t=0}^T,(\hat{X}^i_{0:T})_{i=0}^N, (W^i_T)_{i=0}^N) \\
            \end{array}{}
            \right . \label{EM_HMM_exp_family}
        \end{equation}{}
         
\section{Higher order Neural HMMs} 
    HMMS can also easily incorporate bounded memory processes. Indeed, let $\tau_e, \tau_t\geq 0$ be two memory window sizes, then Equation \eqref{HMM} can be transformed into: 
    \begin{equation}
            \left .
            \begin{array}{ccc}
                x_0  \sim \mu(x) ; &
                y_t \sim p_{g_\theta(x_t, y_{t-\tau_e:t-1})}(y) ; &
                x_{t+1} \sim q_{f_\theta(x_t, y_{t-\tau_t:t})}(x)
            \end{array}
            \right .
        \label{HMM_neural_generalized}
    \end{equation}
    This model can be seen as a neural HMM of order $(\tau_e, \tau_t)$. All the previous theoretical results hold true, in particular Equation \eqref{Q_neural_HMM_} can be used to optimise the parameters of this model by rewriting the loss function from Equation \eqref{loss} as: 
        \begin{equation}
            \left .
            \begin{array}{l}
                l_{\theta_k}(s)(\theta) =  W^{i,j}_T \big[\log\big( p_{g_\theta(\hat{X}^{i,j}_t, y^j_{t-\tau_e:t-1} )}(y_t^j) \big) 
                + \log\big( q_{f_\theta(\hat{X}^{i,j}_{t-1}, y^j_{t-\tau_t:t})}(\hat{X}^{i,j}_t)  \big)\big]
            \end{array}
            \right .
            \label{loss_higher order}
        \end{equation}
    This enables HMMs to piggy-ride on all the progress that Deep Learning is currently undergoing. 
\section{Complementary experimental results}
    \subsection{Synthetic setting: data generating process}
    \label{sec:generatif process}
\begin{figure}[ht]\includegraphics[width=\linewidth]{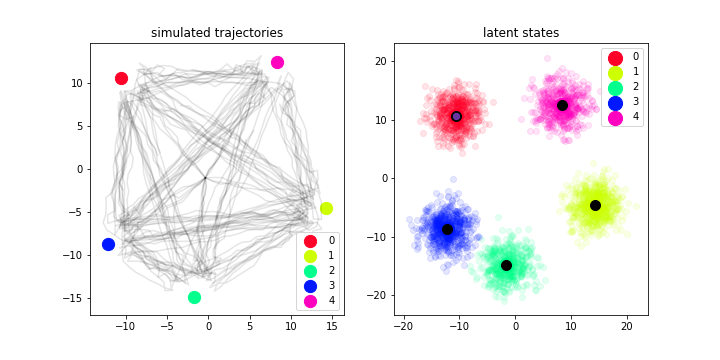}       \caption{Synthetic data set}
        \label{simulations}
        \end{figure}

       We consider a set of targets $(X_i)_1^N$. Particles move toward their current targets $X_{i_t}$:  the direction of the movement is sampled according to a Gaussian $\mathcal{N}(X_{i_t}, \sigma)$. Once a particle is $\epsilon$ close to $X_{i_t}$, its new target is sampled according to a uniform distribution, as modelled by Equation \eqref{synthetic}.
        \begin{equation}
            \left \{
            \begin{array}{cll}
                i_t & = &   \left \{
                            \begin{array}{cl}
                                \sim \mathcal{U}_{[1,2,..,N]} & \textit{if } \; ||y_t - X_{i_{t-1}}|| \leq \epsilon\\
                                i_{t-1} & \textit{if } \; ||y_t - X_{i_{t-1}}|| > \epsilon
                            \end{array}
                            \right . \\
                x_t & \sim &  \mathcal{N}(X_{i_t}, \sigma) \\
                y_{t+1} & = & y_t + \frac{x_t - y_t}{|| x_t - y_t||}
            \end{array}
            \right .
            \label{synthetic}
        \end{equation}{}
        Figure \ref{simulations} illustrates an example of the generated trajectories ($y_t$) and latent states ($x_t$), where the number of components is $N=5$ and the feature dimension is $d=2$.
        
    \subsection{Real-life data sets}
        The real strength of using HMMs is the interpretability of the latent states. We explore two real-life data sets in order to provide qualitative proof that this aspect is not lost when using neural HMMs.
        \subsubsection{Open air museum}
\begin{figure}[ht]  
                \includegraphics[width=.45\linewidth]{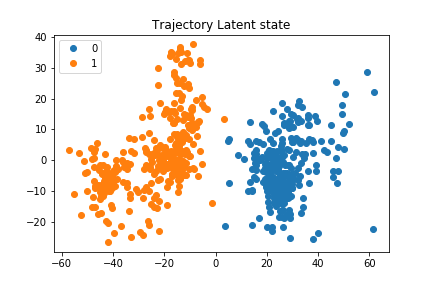}
                \includegraphics[width=.45\linewidth]{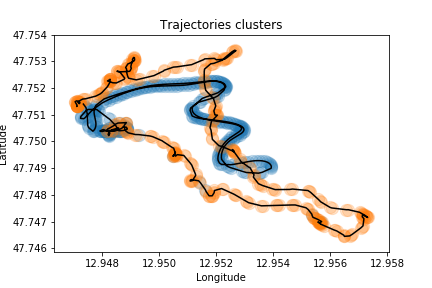}
                \caption{GPS track clusters}        
                \label{museum_clusters}
            \end{figure}

            We consider the data set gathered during a study conducted in an open air museum in Austria~\cite{OpenAirMuseum}. $200$ visitors of the \textit{Freilicht museum} were equipped with GPS-Trackers and their behaviours were qualitatively analysed \cite{OpenAirMuseum}. 
            Given that a visitor's intention and preferences can explain his path when visiting the museum, the HMM model lend itself to explain the observed trajectories. We use the data from the same study and model the trajectories using Equation \eqref{HMM_neural} where $y_t$ is the position of the tourist at time $t$. We use Gaussian distribution for the emission and transition, where the parameters are estimated using 3-layer ANNs. The latent variables are encoded in 2 dimensions.
            Maximising the Log-likelihood of the model using Equation \eqref{Q_neural_HMM_} converges in a few iterations. The parameters are then used to compute the latent states of a given trajectory. Clustering these latent states using K-mean, reveals two visiting behaviours of the museum as illustrated in Figure \ref{museum_clusters}. When reviewing the website of the museum, we discover that the visitors have the choice between discovering the attractions by walking or using a railway ride. The train trajectory can be matched to the blue data point and the walking itinerary to the orange ones. Extracting these patterns in an unsupervised way using only the position of the visitors and despite the fact that the walking and the railway paths intersect, is a good example of how highly interpretable neural HMMs can be.
            
        \subsubsection{Guitar chords}
            Our objective is to provide a prototype music recognition system that relies on neural HMMs, in an unsupervised manner. Let us describe more precisely this example.
            
        \textbf{Setting:}
                We limit the analysis to 10 basic guitar chords (A, Am, Bm, C, D, Dm, E, Em, F, G) for complexity reason, as standard music related research~\cite{guitar_data}. The objective is to train a classifier that decomposes a melody into a  sequence of these chords.
                
                Successful attempts are based on the so-called Pitch Class Profile (PCP)~\cite{guitar_data}. Roughly speaking, the PCP is a set of features obtained through a spectral analysis, followed by a Fourier transform, of the signal, i.e., the melody. Mathematically, it is a mapping from $[0,T]$  to $\mathds{R}^{12}$. Very roughly speaking, it associates to each instant  $t \in [0,T]$ of the melody different energy levels of the 12 keys of a piano octave.

            \textbf{Dataset :}
                A dataset of guitar chords recordings (200 per chord) in various settings (an anechoic chamber / a noisy environment), using four different guitars, and with different playing techniques~\cite{guitar_data} is used. These sequences can be seen as random pieces of music. 
                
                In theory,  each chord's ideal PCP would always activate the same different subsets of harmonics; However, the surrounding noise, the type of guitar, etc.\ are many different reasons that explain actual variations in PCP. In practice, a chord might not be recoverable directly from a single PCP observation: the associated PCP diagram, may vary depending on the player technique, the order in which the different notes of a chord are heard.
                
            \textbf{neuralHMM:}
                The chord successions can be well modeled by 
                higher order neural HMMs, introduced in Equation~\eqref{HMM_neural_generalized}. The observations $y_t$ are the PCP while the latent variables are the chords played.
                
                We recall that a higher order neural HMM is defined by two hyper-parameters (the window memory) and two neural networks.  The experimental results corresponds to
                
                -- Hyper-parameters: $\tau_e = 1$ and $\tau_t = 15$.
                
                -- Emission network $g_\theta$:  a two layer neural network (width of 32 neurons)
                
                -- Transition network $f_\theta$:  an LSTM  constructs an embedding of the previous observations $y_{t-\tau_t-1:t-1}$; they are concatenated with the previous representations and fed to a two layer neural network (same width of 32 neurons).

\begin{figure}[ht]
                \includegraphics[width=\linewidth]{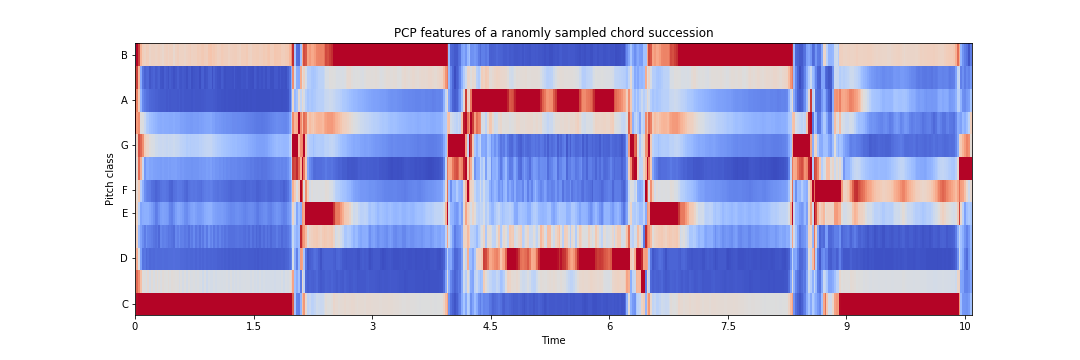}
                \caption{PCP of a randomly sampled guitar chords}
                \label{guitar_chords}   
                \end{figure}

                The optimization of the different neural nets (see Equation~\eqref{Q_neural_HMM_}) require samples of trajectories; for illustration purpose, we fix their lengths to 5 guitar chords (see such a PCP illustrated in Figure~\ref{guitar_chords} corresponding to the sequence ('C', 'E', 'D', 'E', 'F').) and we choose  5 chord recordings  at random in the chord data-set. The total number of possible 'chord trajectories' is prohibitively large, up to $3.10^{16}$, so we either limit its size (for Equation \eqref{Q_neural_HMM_}) or we simply use the  stochastic variant of the algorithm.
                
\begin{figure}[ht]
                \centering
                    \begin{subfigure}{.45\linewidth}
                        \includegraphics[width=\linewidth]{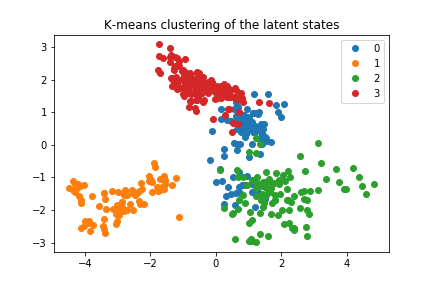}
                        \caption{Latent state embedding clustered using K-mean}
                        \label{latent_guitar}
                    \end{subfigure}
                    \begin{subfigure}{.45\linewidth}
                        \includegraphics[width=\linewidth]{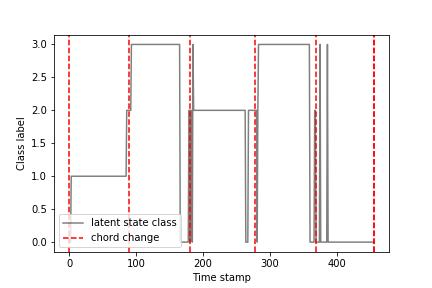}
                        \caption{Class evaluation over time}
                        \label{guitar_seg}
                    \end{subfigure}
                    \caption{Performance analysis of the regretted detection delay as a function of the cutting threshold}
                    \label{perf}
                \end{figure}

            \textbf{Results:}
                Performance wise, we observe in Table \ref{result_table_guitar} that neural HMMs out-performs vanilla recurrent neural networks when using low latent space dimension. In the following, we will use $d_h = 5$. After the training, we  extract the latent states using Algorithm~\ref{bootstrap}. Going back to the example of the trajectory of Figure~\ref{guitar_chords}, we further cluster the associated representations  into four classes (number of unique chords used in this sample) using K-means. This is illustrated in Figure \ref{latent_guitar} after a PCA of the latent states coordinate.
                
               \begin{table}[ht]
                    \caption{Prediction error for various latent space dimensions.}
                    \label{result_table_guitar}
                    \vskip 0.15in
                    \begin{center}
                    \begin{small}
                    \begin{sc}
                    \begin{tabular}{lccccc}
                    \toprule
                    Models & $d_h = 2$ & $d_h = 5$ & $d_h = 10$ & $d_h = 20$  \\
                    \midrule
                    N-HMM & \textbf{.30 $\pm$ .02}& \textbf{.28 $\pm$ .03}& \textbf{.31 $\pm$ .03}& .33 $\pm$ .04 \\
                    HMM       & 1.04 $\pm$ .05 & .65 $\pm$ .03 & .46 $\pm$ .03 & .56 $\pm$ .04 \\
                    RNN       & .64 $\pm$ .07 & .41 $\pm$ .06 & \textbf{.31 $\pm$ .07} & \textbf{.29 $\pm$ .05} \\
                    LSTM       & .62 $\pm$ .06 & .42 $\pm$ .07 & \textbf{.30 $\pm$ .07} & \textbf{.29 $\pm$ .07} \\
                    GRU        & .60 $\pm$ .05 & .40 $\pm$ .07 & \textbf{.31 $\pm$ .08} & .30 $\pm$ .08  \\
                    \bottomrule
                    \end{tabular}
                    \end{sc}
                    \end{small}
                    \end{center}
                    \vskip -0.1in
                \end{table}
                To highlight even more the powerful results of our method, we filter the audio file according to the identified clusters. Figure \ref{guitar_seg}  reveals that each one of them can be associated with a particular chord. Indeed, the dashed red lines are associated with the time step where a new guitar chord is being played, the black line evaluates the cluster label of the audio file over time according to the learned latent states. Those clusters almost coincide.
                
\begin{figure}[ht]
                    \includegraphics[width=\linewidth]{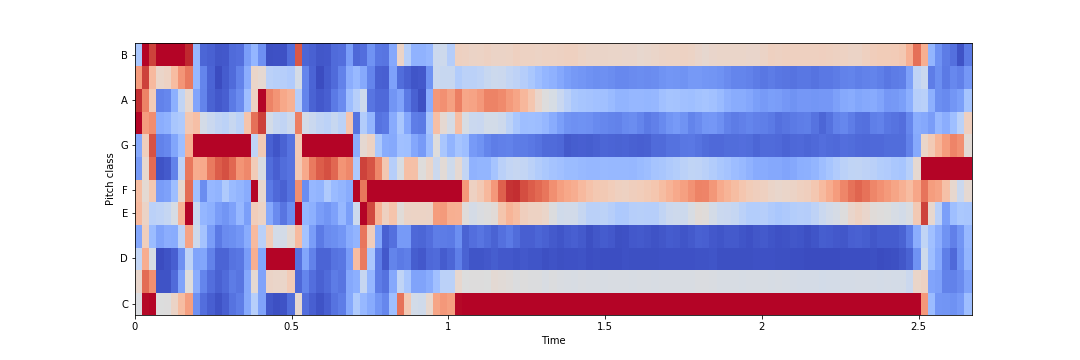}
                    \includegraphics[width=\linewidth]{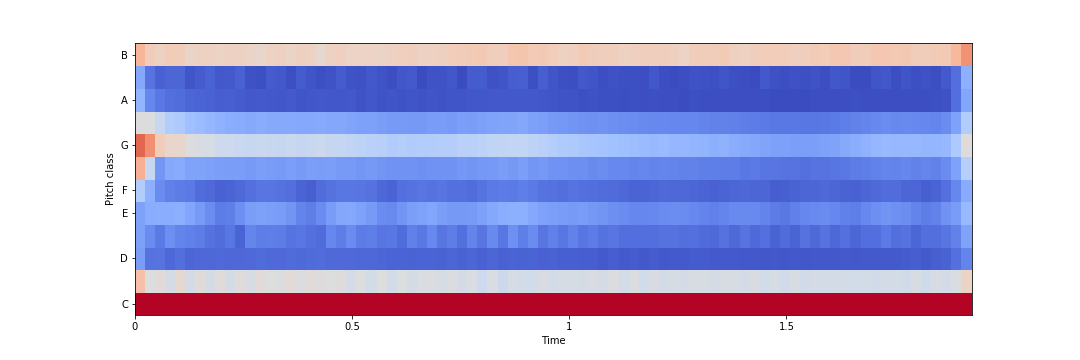}
                    \includegraphics[width=\linewidth]{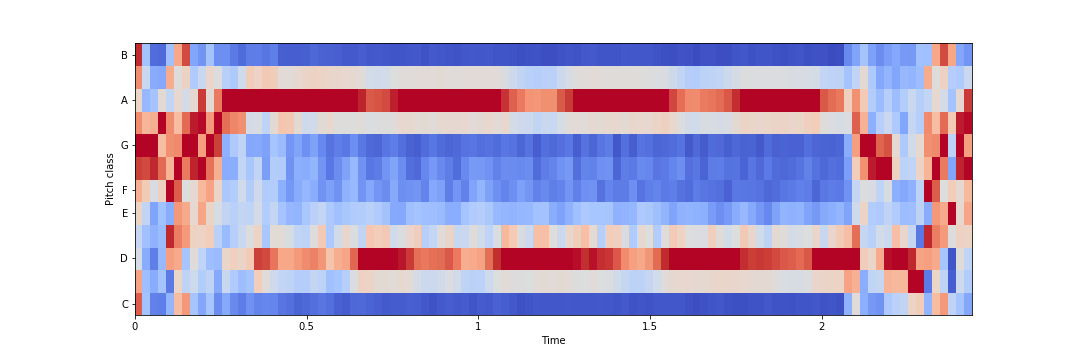}
                    \includegraphics[width=\linewidth]{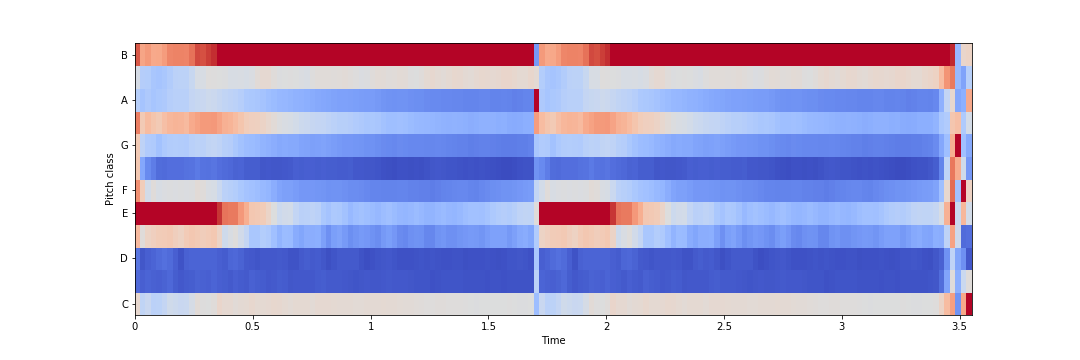}
                    \caption{PCP associated to the latent states classes}
                    \label{guitar_chords_filtered}
                \end{figure}

                Furthermore, we can even reconstruct the typical PCP of a particular chord. We represent them in Figure \ref{guitar_chords_filtered}, where the PCP of the latent spaces in the same specific cluster are concatenated for illustration purposes. 

\end{document}